\documentclass[11pt,a4paper]{article}
\usepackage[hyperref]{acl2018}
\usepackage{times}
\usepackage{latexsym}

\usepackage{graphicx}
\usepackage{booktabs}
\usepackage{amsmath}
\usepackage{todonotes}
\usepackage{multirow}
\usepackage{enumitem}
\usepackage{amsfonts}
\usepackage{url}
\usepackage{setspace}

\renewcommand{\sectionautorefname}{\S\kern-0.2em}
\renewcommand{\subsectionautorefname}{\S\kern-0.2em}
\renewcommand{\subsubsectionautorefname}{\S\kern-0.2em}

\newcommand{\U}{\mathbb{U}}

\aclfinalcopy 

\title{Learning to Ask Good Questions: Ranking Clarification Questions using Neural Expected Value of Perfect Information}

\author{Sudha Rao \\
University of Maryland, College Park \\
{\tt raosudha@cs.umd.edu} \\ \And
Hal Daum\'e III \\
University of Maryland, College Park \\
Microsoft Research, New York City\\
{\tt hal@cs.umd.edu} }

\date{}

\begin{document}
\maketitle

\begin{abstract}
Inquiry is fundamental to communication, and machines cannot effectively collaborate with humans unless they can ask questions. In this work, we build a neural network model for the task of ranking clarification questions. Our model is inspired by the idea of expected value of perfect information: a good question is one whose expected answer will be useful. We study this problem using data from StackExchange, a plentiful online resource in which people routinely ask clarifying questions to posts so that they can better offer assistance to the original poster. We create a dataset of clarification questions consisting of $\sim$77K posts paired with a clarification question (and answer) from three domains of StackExchange: \texttt{\footnotesize askubuntu}, \texttt{\footnotesize unix} and \texttt{\footnotesize superuser}. We evaluate our model on 500 samples of this dataset against expert human judgments and demonstrate significant improvements over controlled baselines.
\end{abstract}

\section{Introduction}\label{introduction}

\begin{figure}[t]
\centering
\setlength\fboxsep{1pt}
\setlength\fboxrule{0.5pt}
\fbox{\includegraphics[width=0.47\textwidth]{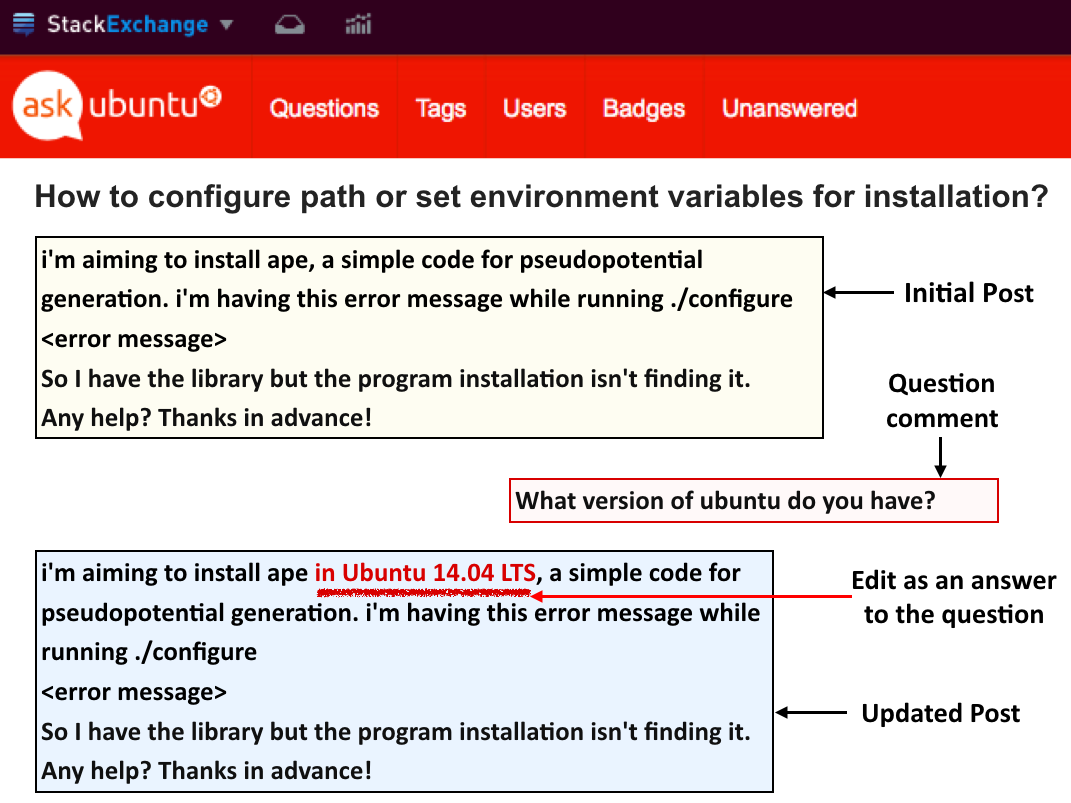}}
\caption{A post on an online Q \& A forum ``askubuntu.com'' is updated to fill the missing information pointed out by the question comment.}
\label{askubuntu_post}
\vspace{-1.1em}
\end{figure}

A principle goal of asking questions is to fill information gaps, typically through clarification questions.\footnote{We define `clarification question' as a question that asks for some information that is currently missing from the given context.} 
We take the perspective that a good question is the one whose \emph{likely answer} will be useful.
Consider the exchange in \autoref{askubuntu_post}, in which an initial poster (who we call ``Terry'') asks for help configuring environment variables.
This post is underspecified and a responder (``Parker'') asks a clarifying question \textsf{\small(a)} below, but could alternatively have asked \textsf{\small(b)} or \textsf{\small(c)}:

\textsf{\small(a) What version of Ubuntu do you have?}

\textsf{\small(b) What is the make of your wifi card?}

\textsf{\small(c) Are you running Ubuntu 14.10 kernel 4.4.0-59-generic on an x86\_64 architecture?}

\noindent
Parker should not ask (b) because an answer is unlikely to be useful; they should not ask (c) because it is too specific and an answer like ``No'' or ``I do not know'' gives little help. 
Parker's question (a) is much better: it is both likely to be useful, and is plausibly answerable by Terry.

In this work, we design a model to rank a candidate set of clarification questions by their usefulness to the given post.
We imagine a use case (more discussion in \autoref{sec:conclusion}) in which, while Terry is writing their post, a system suggests a shortlist of questions asking for information that it thinks people like Parker might need to provide a solution, thus enabling Terry to immediately clarify their post, potentially leading to a much quicker resolution.
Our model is based on the decision theoretic framework of the Expected Value of Perfect Information (EVPI) \cite{avriel1970value}, a measure of the value of gathering additional information. 
In our setting, we use EVPI to calculate which questions are most likely to elicit an answer that would make the post more informative. 

\begin{figure*}[t]
  \centering
  \vspace*{-1em}
\includegraphics[width=0.8\textwidth]{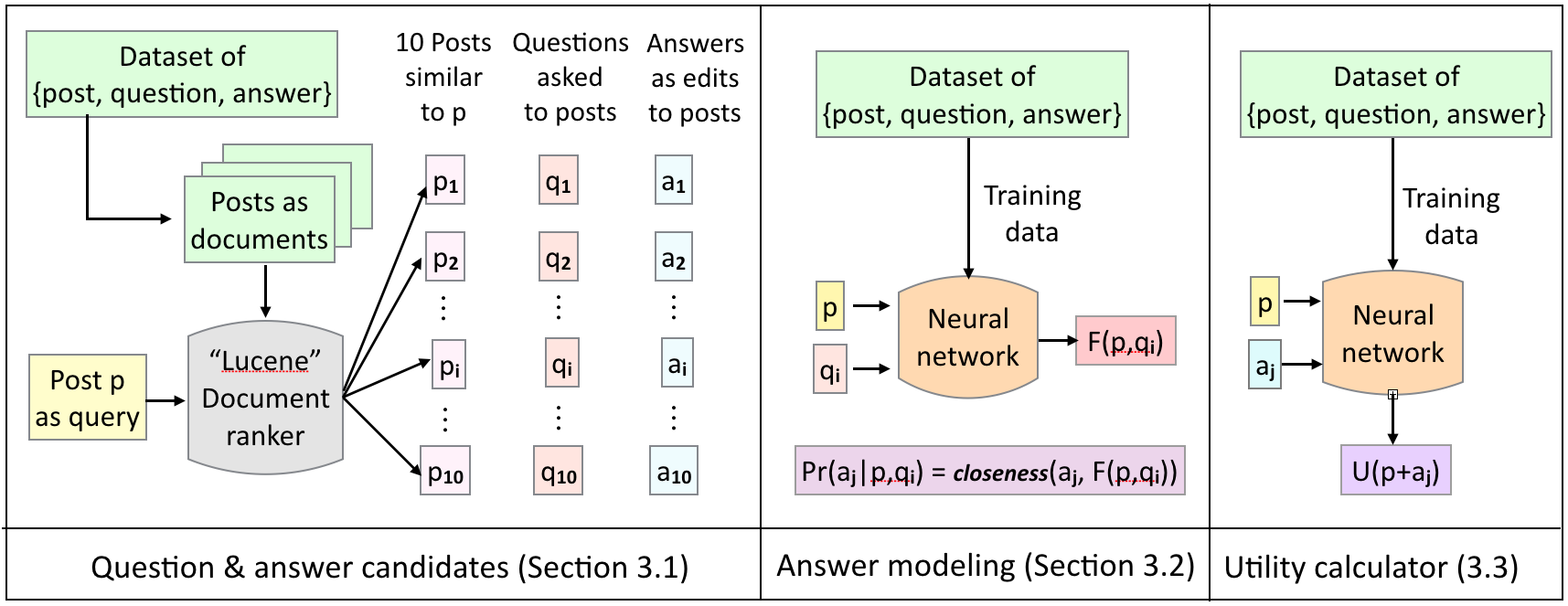}
\caption{ The behavior of our model during test time: Given a post $p$, we retrieve 10 posts similar to post $p$ using Lucene. The questions asked to those 10 posts 
are our question candidates $Q$ and the edits made to the posts in response to the questions are our answer candidates $A$. For each question candidate $q_i$, we generate an answer representation $F(p,q_i)$ and calculate how close is the answer candidate $a_j$ to our answer representation $F(p,q_i)$. We then calculate the utility of the post $p$ if it were updated with the answer $a_j$. Finally, we rank the candidate questions $Q$ by their expected utility given the post $p$ (\autoref{evpi_equation}).}
\label{fig:model}
\end{figure*}

Our work has two main contributions: 
\begin{enumerate}[noitemsep,nolistsep]
\item A novel neural-network model for addressing the task of ranking clarification question built on the framework of expected value of perfect information (\autoref{model}).
\item A novel dataset, derived from StackExchange\footnote{We use data from StackExchange; per license cc-by-sa 3.0, the data is ``intended to be shared and remixed'' (with attribution).}, that enables us to learn a model to ask clarifying questions by looking at the types of questions people ask (\autoref{dataset_creation}).
\end{enumerate}

We formulate this task as a ranking problem on a set of potential clarification questions. We evaluate models both on the task of returning the original clarification question and also on the task of picking any of the candidate clarification questions marked as good by experts (\autoref{evaluation_design}). We find that our EVPI model outperforms the baseline models when evaluated against expert human annotations. We include a few examples of human annotations along with our model performance on them in the supplementary material. We have released our dataset of $\sim$77K $(p,q,a)$ triples and the expert annotations on 500 triples to help facilitate further research in this task.\footnote{\url{https://github.com/raosudha89/ranking_clarification_questions}}


\begin{figure*}[ht]
\centering
\includegraphics[width=0.9\textwidth]{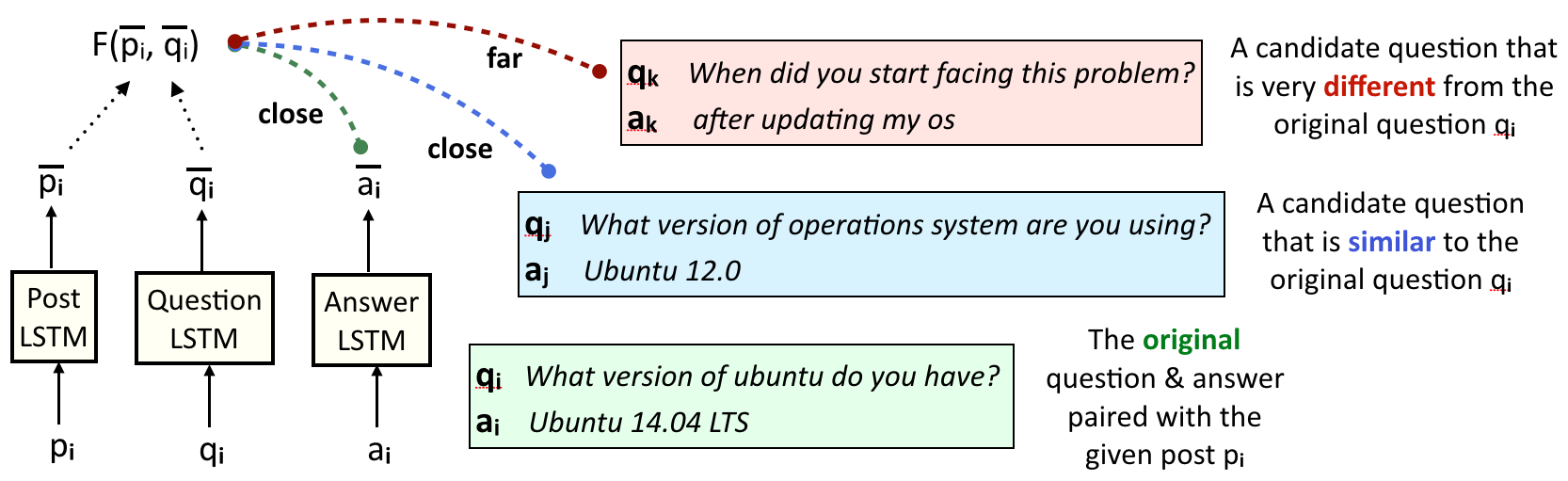}
\caption{Training of our answer generator. Given a post $p_i$ and its question $q_i$, we generate an answer representation that is not only close to its original answer $a_i$, but also close to one of its candidate answers $a_j$ if the candidate question $q_j$ is close to the original question $q_i$.}
\label{fig_answer_generator}
\end{figure*}

\section{Model description}\label{model}

We build a neural network model inspired by the theory of expected value of perfect information (EVPI). EVPI is a measurement of: if I were to acquire information X, how useful would that be to me?  However, because we haven't acquired X yet, we have to take this quantity in expectation over all possible X, weighted by each X's likelihood. In our setting, for any given question $q_i$ that we can ask, there is a set $A$ of possible answers that could be given. For each possible answer $a_j \in A$, there is some probability of getting that answer, and some utility if that were the answer we got. The value of this question $q_i$ is the expected utility, over all possible answers:
\begin{equation}\label{evpi_equation}
 \text{EVPI}(q_i | p) = \sum_{a_j \in A} \mathbb{P}[a_j | p, q_i] \U(p+a_j)
\end{equation} 


In \autoref{evpi_equation}, $p$ is the post, $q_i$ is a potential question from a set of candidate questions $Q$ and $a_j$ is a potential answer from a set of candidate answers $A$. 
Here, $\mathbb{P}[a_j | p, q_i]$ measures the probability of getting an answer $a_j$ given an initial post $p$ and a clarifying question $q_i$, and $\U(p+a_j)$ is a utility function that measures how much more complete $p$ would be if it were augmented with answer $a_j$. 
The modeling question then is how to model: 
\begin{enumerate}[noitemsep,nolistsep]
\item The probability distribution $\mathbb{P}[a_j | p, q_i]$ and
\item The utility function $\U(p+a_j)$.
\end{enumerate}
In our work, we represent both using neural networks over the appropriate inputs. We train the parameters of the two models jointly to minimize a joint loss defined such that an answer that has a higher potential of increasing the utility of a post gets a higher probability.

\autoref{fig:model} describes the behavior of our model during test time. 
Given a post $p$, we generate a set of candidate questions and a set of candidate answers (\autoref{question_candidate_generator}).
Given a post $p$ and a question candidate $q_i$, we calculate how likely is this question to be answered using one of our answer candidates $a_j$ (\autoref{answer_modeling}).
Given a post $p$ and an answer candidate $a_j$, we calculate the utility of the updated post i.e. $\U(p + a_j)$ (\autoref{utility_calculator}).
We compose these modules into a joint neural network that we optimize end-to-end over our data (\autoref{neural_network}).

\subsection{Question \& answer candidate generator}\label{question_candidate_generator}

Given a post $p$, our first step is to generate a set of question and answer candidates. One way that humans learn to ask questions is by looking at how others ask questions in a similar situation. Using this intuition we generate question candidates for a given post by identifying posts similar to the given post and then looking at the questions asked to those posts. For identifying similar posts, we use Lucene\footnote{\url{https://lucene.apache.org/}}, a software extensively used in information retrieval for extracting documents relevant to a given query from a pool of documents. Lucene implements a variant of the term frequency-inverse document frequency (TF-IDF) model to score the extracted documents according to their relevance to the query. We use Lucene to find the top 10 posts most similar to a given post from our dataset (\autoref{dataset_creation}). We consider the questions asked to these 10 posts as our set of question candidates $Q$ and the edits made to the posts in response to the questions as our set of answer candidates $A$. Since the top-most similar candidate extracted by Lucene is always the original post itself, the original question and answer paired with the post is always one of the candidates in $Q$ and $A$. \autoref{dataset_creation} describes in detail the process of extracting the (\textit{post, question, answer}) triples from the StackExchange datadump. 

\subsection{Answer modeling}\label{answer_modeling}

Given a post $p$ and a question candidate $q_i$, our second step is to calculate how likely is this question to be answered using one of our answer candidates $a_j$. We first generate an answer representation by combining the neural representations of the post and the question using a function $F_{ans}(\bar p, \bar q_i)$ (details in \autoref{neural_network}). Given such a representation, we measure the distance between this answer representation and one of the answer candidates $a_j$ using the function below:

\small
\begin{align}
dist(F_{ans}(\bar p, \ \bar q_i), \ \hat a_j) = 1 - cos\_sim(F_{ans}(\bar p, \ \bar q_i), \ \hat a_j) \nonumber
\end{align}
\normalsize

The likelihood of an answer candidate $a_j$ being the answer to a question $q_i$ on post $p$ is finally calculated by combining this distance with the cosine similarity between the question $q_i$ and the question $q_j$ paired with the answer candidate $a_j$: 

\small
\begin{align}\label{eq_answer_prob}
\mathbb{P}[a_j |p ,q_i]  
&=  \exp^{- dist(F_{ans}(\bar p, \ \bar q_i), \ \hat a_j)} * cos\_sim( \hat q_i, \ \hat q_j )
\end{align}
\normalsize

where $\hat a_j$, $\hat q_i$ and $\hat q_j$ are the average word vector of $a_j$, $q_i$ and $q_j$ respectively (details in \autoref{neural_network}) and $cos\_sim$ is the cosine similarity between the two input vectors.

We model our answer generator using the following intuition: a question can be asked in several different ways. For e.g. in \autoref{askubuntu_post}, the question ``\textsf{\small What version of Ubuntu do you have?}'' can be asked in other ways like ``\textsf{\small What version of operating system are you using?}'', ``\textsf{\small Version of OS?}", etc.  
Additionally, for a given post and a question, there can be several different answers to that question. For instance, ``\textsf{\small Ubuntu 14.04 LTS}", ``\textsf{\small Ubuntu 12.0}", ``\textsf{\small Ubuntu 9.0}", are all valid answers. To generate an answer representation capturing these generalizations, we train our answer generator on our triples dataset (\autoref{dataset_creation}) using the loss function below:

\small
\begin{align}\label{eq_answer_generator}
  \textrm{loss}_{\textrm{ans}}(p_i, q_i, a_i, Q_i) 
  &=  { dist(F_{ans}(\bar p_i, \bar q_i), \ \hat a_i}) & \\
  &\hspace{-25mm} +  \sum_{j \in Q} \Big ( {dist(F_{ans}(\bar p_i, \bar q_i), \ \hat a_j)} * cos\_sim( \hat q_i, \ \hat q_j ) \Big ) &\nonumber
\end{align}
\normalsize
where, $\hat a$ and $\hat q$ is the average word vectors of $a$ and $q$ respectively (details in \autoref{neural_network}), $cos\_sim$ is the cosine similarity between the two input vectors.

This loss function can be explained using the example in \autoref{fig_answer_generator}. Question $q_i$ is the question paired with the given post $p_i$. In \autoref{eq_answer_generator}, the first term forces the function $F_{ans}(\bar p_i, \bar q_i)$ to generate an answer representation as close as possible to the correct answer $a_i$. Now, a question can be asked in several different ways. Let $Q_i$ be the set of candidate questions for post $p_i$, retrieved from the dataset using Lucene (\autoref{question_candidate_generator}). Suppose a question candidate $q_j$ is very similar to the correct question $q_i$ ( i.e. $cos\_sim(\hat q_i, \ \hat q_j)$ is near zero). Then the second term forces the answer representation $F_{ans}(\bar p_i, \bar q_i)$ to be close to the answer $a_j$ corresponding to the question $q_j$ as well. Thus in \autoref{fig_answer_generator}, the answer representation will be close to $a_j$ (since $q_j$ is similar to $q_i$), but may not be necessarily close to $a_k$ (since $q_k$ is dissimilar to $q_i$).

\subsection{Utility calculator}\label{utility_calculator}
Given a post $p$ and an answer candidate $a_j$, the third step is to calculate the utility of the updated post i.e. $\U(p + a_j)$. As expressed in \autoref{evpi_equation}, this utility function measures how useful it would be if a given post $p$ were augmented with an answer $a_j$ paired with a different question $q_j$ in the candidate set. Although theoretically, the utility of the updated post can be calculated only using the given post ($p$) and the candidate answer ($a_j$), empirically we find that our neural EVPI model performs better when the candidate question ($q_j$) paired with the candidate answer is a part of the utility function. We attribute this to the fact that much information about whether an answer increases the utility of a post is also contained in the question asked to the post. 
We train our utility calculator using our dataset of $(p, q, a)$ triples (\autoref{dataset_creation}). We label all the $(p_i, q_i, a_i)$ pairs from our triples dataset with label $y=1$. To get negative samples, we make use of the answer candidates generated using Lucene as described in \autoref{question_candidate_generator}. For each $a_j \in A_i$, where $A_i$ is the set of answer candidates for post $p_i$, we label the pair $(p_i, q_j, a_j)$ with label $y=0$, except for when $a_j = a_i$. Thus, for each post $p_i$ in our triples dataset, we have one positive sample and nine negative samples. It should be noted that this is a noisy labelling scheme since a question not paired with the original question in our dataset can often times be a \textit{good} question to ask to the post (\autoref{evaluation_design}). However, since we do not have annotations for such other good questions at train time, we assume such a labelling.

Given a post $p_i$ and an answer $a_j$ paired with the question $q_j$, we combine their neural representations using a function $F_{util}(\bar{p_i}, \bar{q_j}, \bar{a_j})$ (details in \autoref{neural_network}). 
The utility of the updated post is then defined as $\U(p_i + a_j) = \sigma ( F_{util}(\bar{p_i}, \bar{q_j}, \bar{a_j}) )$\footnote{$\sigma$ is the sigmoid function.}. We want this utility to be close to 1 for all the positively labelled $(p,q,a)$ triples and close to 0 for all the negatively labelled $(p, q, a)$ triples. We therefore define our loss using the binary cross-entropy formulation below:
\begin{align}\label{eq_utility_calculator}
  \textrm{loss}_{\textrm{util}}(y_i, \bar p_i, \bar q_j, \bar a_j) &= y_i \log(\sigma (F_{util}(\bar{p_i}, \bar{q_j}, \bar{a_j})))
\end{align}

\subsection{Our joint neural network model}\label{neural_network}
Our fundamental representation is based on recurrent neural networks over word embeddings. We obtain the word embeddings using the GloVe \cite{pennington2014glove} model trained on the entire datadump of StackExchange.\footnote{Details in the supplementary material.}. In \autoref{eq_answer_prob} and \autoref{eq_answer_generator}, the average word vector representations $\hat q$ and $\hat a$ are obtained by averaging the GloVe word embeddings for all words in the question and the answer respectively. Given an initial post $p$, we generate a post neural representation $\bar{p}$ using a post LSTM (long short-term memory architecture) \cite{hochreiter1997long}. The input layer consists of word embeddings of the words in the post which is fed into a single hidden layer. The output of each of the hidden states is averaged together to get our neural representation $\bar p$.
Similarly, given a question $q$ and an answer $a$, we generate the neural representations $\bar{q}$ and $\bar{a}$ using a question LSTM and an answer LSTM respectively. We define the function $F_{ans}$ in our answer model as a feedforward neural network with five hidden layers on the inputs $\bar{p}$ and $\bar{q}$. Likewise, we define the function $F_{util}$ in our utility calculator as a feedforward neural network with five hidden layers on the inputs $\bar{p}$, $\bar{q}$ and $\bar{a}$. We train the parameters of the three LSTMs corresponding to $p$, $q$ and $a$, and the parameters of the two feedforward neural networks jointly to minimize the sum of the loss of our answer model (\autoref{eq_answer_generator}) and our utility calculator (\autoref{eq_utility_calculator}) over our entire dataset:
\begin{align}
                       \sum_i \sum_j \textrm{loss}_{\textrm{ans}}(\bar p_i, \bar q_i, \bar a_i, Q_i)  
                        +  \textrm{loss}_{\textrm{util}}(y_i, \bar p_i, \bar q_j, \bar a_j)
\end{align}

Given such an estimate $\mathbb{P}[a_j|p, q_i]$ of an answer and a utility $\U(p+a_j)$ of the updated post, 
we rank the candidate questions by their value as calculated using \autoref{evpi_equation}. 
The remaining question, then, is how to get data that enables us to train our answer model and our utility calculator. Given data, the training becomes a multitask learning problem, where we learn simultaneously to predict utility and to estimate the probability of answers.


\section{Dataset creation}\label{dataset_creation}
StackExchange is a network of online question answering websites about varied topics like academia, ubuntu operating system, latex, etc. 
The data dump of StackExchange contains timestamped information about the posts, comments on the post and the history of the revisions made to the post. We use this data dump to create our dataset of (\textit{post, question, answer}) triples: where the \textit{post} is the initial unedited post, the \textit{question} is the comment containing a question and the \textit{answer} is either the edit made to the post after the question or the author's response to the question in the comments section.

\paragraph{Extract posts:} We use the post histories to identify posts that have been updated by its author. We use the timestamp information to retrieve the initial unedited version of the post.

\paragraph{Extract questions:} For each such initial version of the post, we use the timestamp information of its comments to identify the first question comment made to the post. We truncate the comment till its question mark '?' to retrieve the question part of the comment. We find that about 7\% of these are rhetoric questions that indirectly suggest a solution to the post. For e.g. \textit{``have you considered installing X?''}. We do a manual analysis of these non-clarification questions and hand-crafted a few rules to remove them. \footnote{Details in the supplementary material.}

\paragraph{Extract answers:} We extract the answer to a clarification question in the following two ways:\\
\textit{(a) Edited post}: Authors tend to respond to a clarification question by editing their original post and adding the missing information. In order to account for edits made for other reasons like stylistic updates and grammatical corrections, we consider only those edits that are longer than four words. Authors can make multiple edits to a post in response to multiple clarification questions.\footnote{On analysis, we find that 35\%-40\% of the posts get asked multiple clarification questions. We include only the first clarification question to a post in our dataset since identifying if the following questions are clarifications or a part of a dialogue is non-trivial.} To identify the edit made corresponding to the given question comment, we choose the edit closest in time following the question.\\
\textit{(b) Response to the question}: Authors also respond to clarification questions as subsequent comments in the comment section. We extract the first comment by the author following the clarification question as the answer to the question.
 
In cases where both the methods above yield an answer, we pick the one that is the most semantically similar to the question, where the measure of similarity is the cosine distance between the average word embeddings of the question and the answer.

We extract a total of 77,097 (\textit{post, question, answer}) triples across three domains in StackExchange (\autoref{data_statistics}). 
We will release this dataset along with the the nine question and answer candidates per triple that we generate using lucene (\autoref{question_candidate_generator}).
We include an analysis of our dataset in the supplementary material.

\begin{table}
\centering
\footnotesize
\begin{tabular}{lccc}
\toprule
& Train & Tune & Test  \\
\midrule
\texttt{\footnotesize askubuntu} & 19,944 & 2493 & 2493\\
\texttt{\footnotesize unix} & 10,882 & 1360 & 1360 \\
\texttt{\footnotesize superuser} & 30,852 & 3857 & 3856\\

\bottomrule
\end{tabular}
\caption{Table above shows the sizes of the train, tune and test split of our dataset for three domains.}\label{data_statistics}
\end{table}

\section{Evaluation design}\label{evaluation_design}
We define our task as given a post $p$, and a set of candidate clarification questions $Q$, rank the questions according to their usefulness to the post. Since the candidate set includes the original question $q$ that was asked to the post $p$, one possible approach to evaluation would be to look at how often the original question is ranked higher up in the ranking predicted by a model. However, there are two problems to this approach: 1) Our dataset creation process is noisy. The original question paired with the post may not be a useful question. For e.g. \textit{``are you seriously asking this question?''}, \textit{``do you mind making that an answer?''}\footnote{Data analysis included in the supplementary material suggests ~9\% of the questions are not useful.}. 2) The nine other questions in the candidate set are obtained by looking at questions asked to posts that are similar to the given post.\footnote{Note that this setting is different from the distractor-based setting popularly used in dialogue \cite{lowe2015ubuntu} where the distractor candidates are chosen randomly from the corpus.}  This greatly increases the possibility of some other question(s) being more useful than the original question paired with the post. This motivates an evaluation design that does not rely solely on the original question but also uses human judgments. We randomly choose a total of 500 examples from the test sets of the three domains proportional to their train set sizes (\texttt{\footnotesize askubuntu:160}, \texttt{\footnotesize unix:90} and \texttt{\footnotesize superuser:250}) to construct our evaluation set.

\subsection{Annotation scheme}\label{annotation_scheme}
Due to the technical nature of the posts in our dataset, identifying useful questions requires technical experts. We recruit 10 such experts on Upwork\footnote{https://upwork.com} who have prior experience in unix based operating system administration.\footnote{Details in the supplementary material.}
We provide the annotators with a post and a randomized list of the ten question candidates obtained using Lucene (\autoref{question_candidate_generator}) and ask them to select a \emph{single} ``best'' ($B$) question to ask, and additionally mark as ``valid'' ($V$) other questions that they thought would be okay to ask in the context of the original post. We enforce that the ``best'' question be always marked as a ``valid'' question. We group the 10 annotators into 5 pairs and assign the same 100 examples to the two annotators in a pair.

\begin{figure}
\includegraphics[scale=0.3]{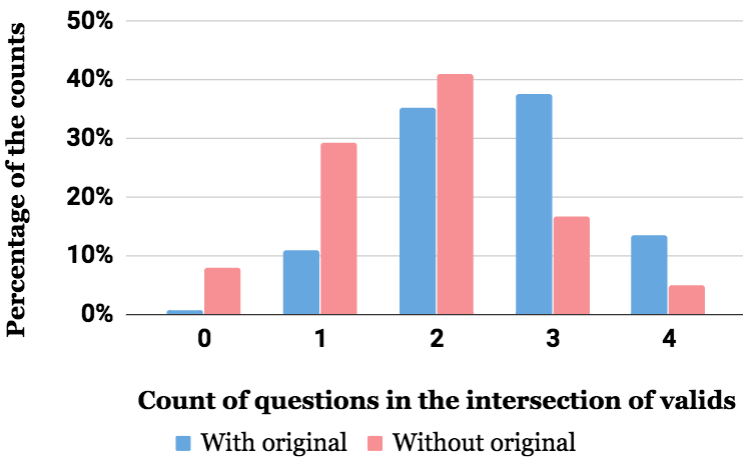}
\caption{\small Distribution of the count of questions in the intersection of the ``valid'' annotations.}\label{fig:valid-dist}
\end{figure}

\subsection{Annotation analysis}\label{annotation_analysis}

We calculate the inter-annotator agreement on the ``best'' and the ``valid'' annotations using Cohen's Kappa measurement. When calculating the agreement on the ``best'' in the strict sense, we get a low agreement of 0.15.  However, when we relax this to a case where the question marked as``best'' by one annotator is marked as ``valid'' by another, we get an agreement of 0.87.
The agreement on the ``valid'' annotations, on the other hand, was higher: 0.58. We calculate this agreement on the binary judgment of whether a question was marked as valid by the annotator.

Given these annotations, we calculate how often is the original question marked as ``best'' or ``valid'' by the two annotators. We find that 72\% of the time one of the annotators mark the original as the ``best'', whereas only 20\% of the time both annotators mark it as the ``best'' suggesting against an evaluation solely based on the original question. 
On the other hand, 88\% of the time one of the two annotators mark it as a ``valid'' question confirming the noise in our training data.\footnote{76\% of the time both the annotators mark it as a ``valid''.}  
 
\autoref{fig:valid-dist} shows the distribution of the counts of questions in the intersection of ``valid'' annotations (blue legend). We see that about 85\% of the posts have more than 2 valid questions and 50\% have more than 3 valid questions. The figure also shows the distribution of the counts when the original question is removed from the intersection (red legend). Even in this set, we find that about 60\% of the posts have more than two valid questions. These numbers  suggests that the candidate set of questions retrieved using Lucene (\autoref{question_candidate_generator}) very often contains useful clarification questions. 

\begin{table*}
\footnotesize
\centering
\begin{tabular}{l | c c c c | c c c c | c}
\toprule
 & \multicolumn{4}{c |}{$B1 \cup B2$} & \multicolumn{4}{c}{$V1 \cap V2$} & \multicolumn{1}{|c}{Original} \\
Model & p@1 & p@3 & p@5 & MAP & p@1 & p@3 & p@5 & MAP & p@1\\
\midrule
Random & 17.5 & 17.5 & 17.5 & 35.2 & 26.4 & 26.4 & 26.4 & 42.1 & 10.0\\
Bag-of-ngrams & 19.4 & 19.4 & 18.7 & 34.4 & 25.6  & 27.6 & 27.5 & 42.7 & 10.7\\
Community QA & 23.1 & 21.2 & 20.0 & 40.2 & 33.6 & 30.8 & 29.1 & 47.0 & 18.5  \\
\midrule
Neural $(p,q)$ & 21.9 & 20.9 & 19.5 & 39.2 & 31.6 & 30.0 & 28.9 & 45.5 & 15.4\\
Neural $(p,a)$ & 24.1 & \bf 23.5 & 20.6 & 41.4 & 32.3 & \bf 31.5 & 29.0 & 46.5 & 18.8\\
Neural $(p,q,a)$ & 25.2 & \bf 22.7 & \bf 21.3 & 42.5 & 34.4 & \bf 31.8 & \bf 30.1 & 47.7 & 20.5\\
\midrule
EVPI & \bf 27.7 & \bf 23.4 & \bf 21.5 & \bf 43.6 & \bf 36.1 & \bf 32.2 & \bf 30.5 & \bf 49.2 & \bf 21.4\\
\bottomrule
\end{tabular}
\caption{ Model performances on 500 samples when evaluated against the union of  the ``best'' annotations ($B1 \cup B2$), intersection of the ``valid'' annotations ($V1 \cap V2$) and the original question paired with the post in the dataset. The difference between the bold and the non-bold numbers is statistically significant with $p < 0.05$ as calculated using bootstrap test. p@k is the precision of the  k questions ranked highest by the model and MAP is the mean average precision of the ranking predicted by the model.}\label{results_on10}
\end{table*}

\section{Experimental results}\label{experiments_results}

Our primary research questions that we evaluate experimentally are:
\begin{enumerate}[noitemsep,nolistsep]
\item Does a neural network architecture improve upon non-neural baselines? 
\item Does the EVPI formalism provide leverage over a similarly expressive feedforward network?
\item Are answers useful in identifying the right question?
\item How do the models perform when evaluated on the candidate questions excluding the original? 
\end{enumerate}

\subsection{Baseline methods}\label{baselines}

We compare our model with following baselines:

\paragraph{Random:} Given a post, we randomly permute its set of 10 candidate questions uniformly.\footnote{We take the average over 1000 random permutations.}

\paragraph{Bag-of-ngrams:} Given a post and a set of 10 question and answer candidates, we construct a bag-of-ngrams representation for the post, question and answer. We train the baseline on all the positive and negative candidate triples (same as in our utility calculator (\autoref{utility_calculator})) to minimize hinge loss on misclassification error using cross-product features between each of (\textit{p, q}), (\textit{q, a}) and (\textit{p, a}). We tune the ngram length and choose n=3 which performs best on the tune set. The question candidates are finally ranked according to their predictions for the positive label.

\paragraph{Community QA:} The recent SemEval2017 Community Question-Answering (CQA) \cite{nakov2017semeval} included a subtask for  ranking a set of comments according to their relevance to a given post in the Qatar Living\footnote{http://www.qatarliving.com/forum} forum. \newcite{nandi2017iit}, winners of this subtask, developed a logistic regression model using features based on string similarity, word embeddings, etc. We train this model on all the positively and negatively labelled $(p, q)$ pairs in our dataset (same as in our utility calculator (\autoref{utility_calculator}), but without $a$). We use a subset of their features relevant to our task.\footnote{Details in the supplementary material.}

\paragraph{Neural baselines: } We construct the following neural baselines based on the LSTM representation of their inputs (as described in \autoref{neural_network}):\\
1. \textbf{Neural(p, q):} Input is concatenation of $\bar{p}$ and $\bar{q}$.\\
2. \textbf{Neural(p, a):} Input is concatenation of $\bar{p}$ and $\bar{a}$.\\
3. \textbf{Neural(p, q, a):} Input is concatenation of $\bar{p}$, $\bar{q}$ and $\bar{a}$.\\
      
Given these inputs, we construct a fully connected feedforward neural network with 10 hidden layers and train it to minimize the binary cross entropy across all positive and negative candidate triples (same as in our utility calculator (\autoref{utility_calculator})).
The major difference between the neural baselines and our EVPI model is in the loss function: the EVPI model is trained to minimize the joint loss between the answer model (defined on $F_{ans}(p, q)$ in \autoref{eq_answer_generator}) and the utility calculator (defined on $F_{util}(p,q,a)$ in \autoref{eq_utility_calculator}) whereas the neural baselines are trained to minimize the loss directly on $F(p,q)$, $F(p,a)$ or $F(p, q, a)$. 
We include the implementation details of all our neural models in the supplementary material. 

\subsection{Results}\label{results}

\subsubsection{Evaluating against expert annotations}\label{using_human_judgments}
We first describe the results of the different models when evaluated against the expert annotations we collect on 500 samples (\autoref{evaluation_design}). Since the annotators had a low agreement on a single best, we evaluate against the union of the ``best'' annotations ($B1 \cup B2$ in \autoref{results_on10}) and against the intersection of the ``valid'' annotations ($V1 \cap V2$ in \autoref{results_on10}).

Among non-neural baselines, we find that the bag-of-ngrams baseline performs slightly better than random but worse than all the other models. The Community QA baseline, on the other hand, performs better than the neural baseline (Neural $(p,q)$), both of which are trained without using the answers.
The neural baselines with answers (Neural$(p,q,a)$ and Neural$(p,a)$) outperform the neural baseline without answers (Neural$(p,q)$), showing that answer helps in selecting the right question. 

More importantly, EVPI outperforms the Neural $(p, q, a)$ baseline across most metrics. Both models use the same information regarding the true question and answer and are trained using the same number of model parameters.\footnote{We use 10 hidden layers in the feedforward network of the neural baseline and five hidden layers each in the two feedforward networks $F_{ans}$ and $F_{util}$ of the EVPI model.}
However, the EVPI model, unlike the neural baseline, additionally makes use of alternate question and answer candidates to compute its loss function. This shows that when the candidate set consists of questions similar to the original question, summing over their utilities gives us a boost.

\subsubsection{Evaluating against the original question}\label{using_original_question}
The last column in \autoref{results_on10} shows the results when evaluated against the original question paired with the post. The bag-of-ngrams baseline performs similar to random, unlike when evaluated against human judgments. The Community QA baseline again outperforms Neural$(p, q)$ model and comes very close to the Neural $(p,a)$ model. 

As before, the neural baselines that make use of the answer outperform the one that does not use the answer and the EVPI model performs significantly better than Neural($p,q,a$).
  
\subsubsection{Excluding the original question}

In the preceding analysis, we considered a setting in which the ``ground truth'' original question was in the candidate set $Q$. While this is a common evaluation framework in dialog response selection \cite{lowe2015ubuntu}, it is overly optimistic. 
We, therefore, evaluate against the ``best'' and the ``valid'' annotations on the nine other question candidates. 
We find that the neural models beat the non-neural baselines. However, the differences between all the neural models are statistically insignificant.\footnote{Results included in the supplementary material.} 

\section{Related work} \label{related_work}

Most prior work on question generation has focused on generating reading comprehension questions:  given text, write questions that one might find on a standardized test \cite{vanderwende2008importance,heilman2011automatic,rus2011question,olney2012question}.  Comprehension questions, by definition, are answerable from the provided text. Clarification questions--our interest--are not.  

Outside reading comprehension questions, \newcite{labutov2015deep} generate high-level question templates by crowdsourcing 
which leads to significantly less data than we collect using our method. \newcite{liu2010automatic} use template question generation to help authors write better related work sections. \newcite{mostafazadeh2016generating} introduce a Visual Question Generation task where the goal is to generate natural questions that are not about what is present in the image rather about what can be inferred given the image, somewhat analogous to clarification questions. 
\newcite{penas2010filling} identify the notion of missing information similar to us, but they fill the knowledge gaps in a text with the help of external knowledge bases, whereas we instead ask clarification questions. 
\newcite{artzi2011bootstrapping} use human-generated clarification questions to drive a semantic parser where the clarification questions are aimed towards simplifying a user query; whereas we generate clarification questions aimed at  identifying missing information in a text. 

Among works that use community question answer forums, the keywords to questions (K2Q) system \cite{zheng2011k2q} generates a list of candidate questions and refinement words, given a set of input keywords, to help a user ask a better question. \newcite{figueroa2013learning} rank different paraphrases of query for effective search on forums. \cite{romeo2016neural} develop a neural network based model for ranking questions on forums with the intent of retrieving similar other question. The recent SemEval-2017 Community Question-Answering (CQA) \cite{nakov2017semeval} task included a subtask to rank the comments according to their relevance to the post. Our task primarily differs from this task in that we want to identify a question comment which is not only relevant to the post but will also elicit useful information missing from the post. \newcite{hoogeveen2015cqadupstack}  created the CQADupStack dataset using StackExchange forums for the task of duplicate question retrieval. Our dataset, on the other hand, is designed for the task of ranking clarification questions asked as comments to a post.


\section{Conclusion} \label{sec:conclusion}

We have constructed a new dataset for learning to rank clarification questions, and proposed a novel model for solving this task.
Our model integrates well-known deep network architectures with the classic notion of expected value of perfect information, which effectively models a pragmatic choice on the part of the questioner: how do I \emph{imagine} the other party would answer if I were to ask this question. Such pragmatic principles have recently been shown to be useful in other tasks as well \cite{golland2010game,smith2013learning,orita2015discourse,andreas2016reasoning}. 
One can naturally extend our EVPI approach to a full reinforcement learning approach to handle multi-turn conversations.

Our results shows that the EVPI model is a promising formalism for the question generation task. In order to move to a full system that can help users like Terry write better posts, there are three interesting lines of future work.
First, we need it to be able to generalize: for instance by constructing templates of the form ``What version of \_\_\_ are you running?'' into which the system would need to fill a variable. 
Second, in order to move from question ranking to question generation, one could consider sequence-to-sequence based neural network models that have recently proven to be effective for several language generation tasks \cite{sutskever2014sequence,serban2016building,yin2016neural}.
Third is in evaluation: given that this task requires expert human annotations and also given that there are multiple possible good questions to ask, how can we automatically measure performance at this task?, a question faced in dialog and generation more broadly \cite{paek2001empirical,lowe2015ubuntu,liu2016not}.

\section*{Acknowledgments}

The authors thank the three anonymous reviewers of this paper, and the anonymous reviewers of the previous versions for their helpful comments and suggestions. They also thank the members of the Computational Linguistics and Information Processing (CLIP) lab at University of Maryland for helpful discussions.  

This work was supported by NSF grant IIS-1618193. Any opinions, findings, conclusions, or recommendations expressed here are those of the authors and do not necessarily reflect the view of the sponsors.

\bibliography{question_generation}
\bibliographystyle{acl_natbib}

\end{document}